\documentclass[journal]{IEEEtran}
\usepackage{graphicx}
\usepackage{booktabs}
\usepackage{amssymb}

\usepackage{algorithm}
\usepackage{algorithmicx}
\usepackage{algpseudocode}
\usepackage{amsmath}
\usepackage{subfigure}
\usepackage{color} %2020/03/24

\usepackage{xspace}

\makeatletter
\DeclareRobustCommand\onedot{\futurelet\@let@token\@onedot}
\def\@onedot{\ifx\@let@token.\else.\null\fi\xspace}

\def\eg{\emph{e.g}\onedot} 
\def\ie{\emph{i.e}\onedot} 
 
\def\etc{\emph{etc}\onedot}

\makeatother

\ifCLASSINFOpdf
\else
\fi
\begin{document}
\title{Where to Look and How to Describe: \\ Fashion Image Retrieval with an Attentional Heterogeneous Bilinear Network}

\author{Haibo~Su,
        Peng~Wang,
        Lingqiao~Liu,
        Hui~Li,
        Zhen~Li,
        and~Yanning~Zhang% <-this % stops a space
\thanks{H. Su, P. Wang and Y. Zhang are with the School of Computer Science, Northwestern Polytechnical University, China, and the National Engineering Laboratory for Integrated Aero-Space-Ground-Ocean Big Data Application Technology, China.
H. Li and L. Liu are with the School of Computer Science, University of Adelaide, Australia. Z. Li is with the MinSheng FinTech Corp. Ltd, China.
P. Wang is the corresponding author (E-mail: peng.wang@nwpu.edu.cn).
}% <-this % stops a space
%\thanks{}% <-this % stops a space
%\thanks{\textcolor{black}{Z. Li is with the MinSheng FinTech Corp. Ltd.}}
}
\markboth{IEEE TRANSACTIONS ON CIRCUITS AND SYSTEMS FOR VIDEO TECHNOLOGY}%
{Shell \MakeLowercase{\textit{et al.}}: Bare Demo of IEEEtran.cls for IEEE Journals}



\maketitle

\begin{abstract}
Fashion products typically feature in compositions of a variety of styles at different clothing parts.
In order to distinguish images of different fashion products, 
we need to extract both appearance (\ie, ``how to describe'') and localization (\ie, ``where to look'') information, and their interactions.  
To this end, we propose a biologically inspired framework for image-based fashion product retrieval, which mimics the hypothesized two-stream visual  processing  system  of  human  brain.
The proposed attentional heterogeneous bilinear network (AHBN) consists of two branches: 
a deep CNN branch to extract fine-grained appearance attributes and a fully convolutional branch to extract landmark localization information.
A joint channel-wise attention mechanism is further applied to the extracted heterogeneous features to focus on important channels, 
followed by a compact bilinear pooling layer to model the interaction of the two streams.
Our proposed framework achieves satisfactory performance on three image-based fashion product retrieval benchmarks.
\end{abstract}


% \fi
\IEEEpeerreviewmaketitle

\section{Introduction}
\IEEEPARstart{I}{mage-based} fashion product retrieval is an effective way of helping customers to browse and search from a vast amount of fashion products. It has a significant commercial value and gains extensive research interest in recent years.  Unlike generic objects, fashion products usually share a lot of appearance similarities and the differences between products can be subtle, \eg, the different styles of necklines such as crew neck, V-neck and boat neck. On the other hand, the visual appearance of the same product may undergo large appearance variations due to background and illumination change as well as pose and perspective differences.

\begin{figure}[t]
\centering
\resizebox{0.9\columnwidth}{!}{
\includegraphics[width=\linewidth]{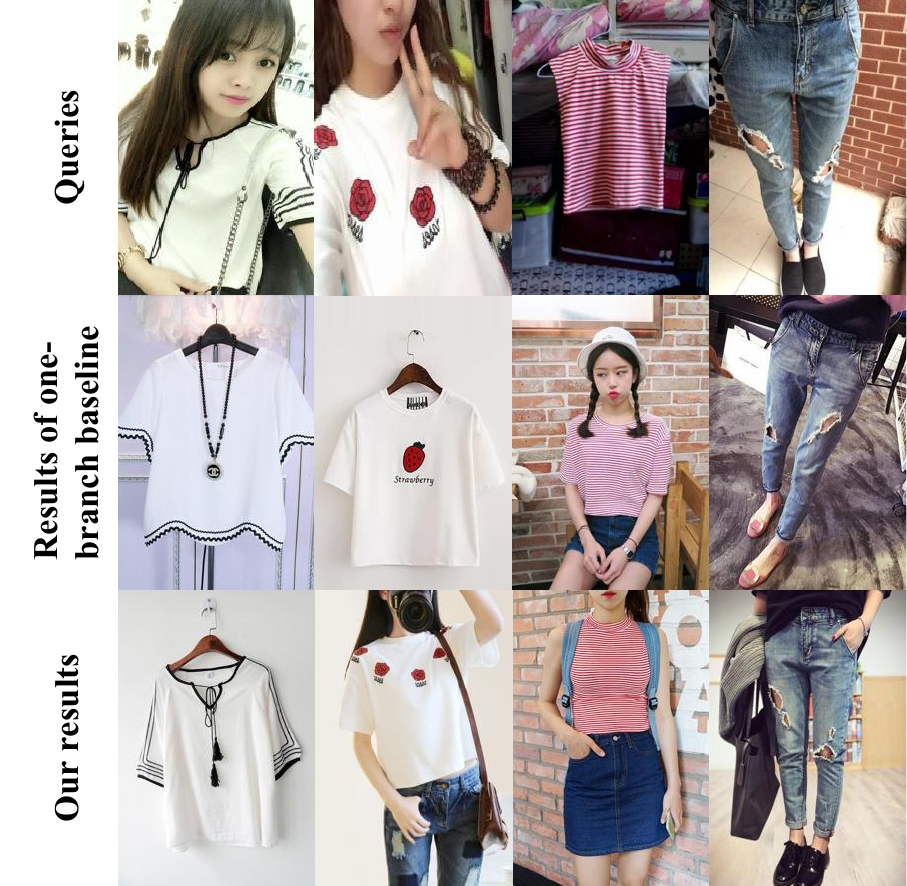}
}
\caption{Illustration of retrieval results. The three rows from top to bottom respectively correspond to 
query images, results of a one-branch strong baseline and results of our heterogeneous two-branch model.
The one-branch baseline makes mistakes when two different items have the same visual attribute at different locations.  
Our model performs better, as it not only extracts visual attributes, but also encodes their locations.}
\label{fig:sample_front}
\end{figure}

These difficulties can be summarized into two issues:
(1) where to look and (2) how to describe. The former issue reflects the challenge of identifying the key parts of an object. A product image usually involves multiple object parts, \eg, sleeves or belt, and the comparison between two product images can be done via comparing the visual appearances of multiple parts. Localizing the object parts and performing the part-level comparison can be beneficial. This is because fashion products are usually articulated objects and localizing part somehow normalizes the visual appearance of images and accounts for the pose variations. In addition, the discrepancy between two similar product images can reside in one or a few key regions, and local comparison on identified parts reduces the difficulty in discerning the subtle differences. The second issue is to obtain a robust descriptor to describe the visual content of product images. Note that the fashion product may have a significant appearance variance due to the change of pose, lighting conditions, \etc. An ideal descriptor should be robust to those variations, but be sensitive to the attribute aspects of a fashion product, \eg, the type of sleeves.

This paper proposes an Attentional Heterogeneous Bilinear Network (AHBN) to simultaneously address the aforementioned two issues. The proposed network has two dedicated branches, one for providing part location information and the other for providing attribute-level descriptors. The outputs from the two branches are then integrated by an attentional bilinear module to generate the image-level representation. The two branches are pre-trained with two auxiliary tasks to ensure the two branches have the capabilities of part localization and attribute description. Specifically, for the first branch, we adopt the hour-glass network and associate it with a landmark prediction task; for the second branch, we adopt the Inception-ResNet-v2 network~\cite{szegedy2017inception} and associate it with an attribute prediction task. The annotations for both tasks are available from the existing dataset and the feature representations from the two branches are employed for creating the image-level representation. Each channel of the feature representations from the two branches might not be equally important. To weight the importance of different channels, we apply a channel-wise attention module for the features from both branches. This attention module is jointly driven by the information from both the part localization branch and the attribute-level description branch. The weighted features are then integrated by using compact bilinear pooling. By evaluating the proposed approach on two large datasets, \eg, DeepFashion dataset~\cite{liu2016deepfashion} and Exact Street2Shop dataset~\cite{hadi2015buy}, we demonstrate that the proposed AHBN can achieve satisfactory retrieval performance and we also validate the benefits of our dual-branch design and proposed attention mechanism. To sum up, our main contributions are as follows:
\begin{itemize}
\item A heterogeneous two-branch design and multi-task training scheme for solving ``where to look'' and ``how to describe'' issues. 
Compared to the homogeneous two-branch design (\eg, \cite{gao2016compact}),  
our heterogeneous model is biologically inspired: it behaves more like the hypothesized two-stream visual processing system of human brain~\cite{goodale1992separate} 
that performs identification and localization in two pathways respectively.
\item An attentional bilinear network for integrating information from the two branches and modeling their pairwise interactions.
A novel channel-wise co-attention module is proposed to mutually guide the generation of channel weights for both branches.
\item Through experimental study, we validate the contribution of the proposed components by its superior performance. 
Our AHBN achieves satisfactory performance on all the three evaluated fashion retrieval benchmarks.
\end{itemize}

\begin{figure*}[t]
\centering
\includegraphics[width=\linewidth]{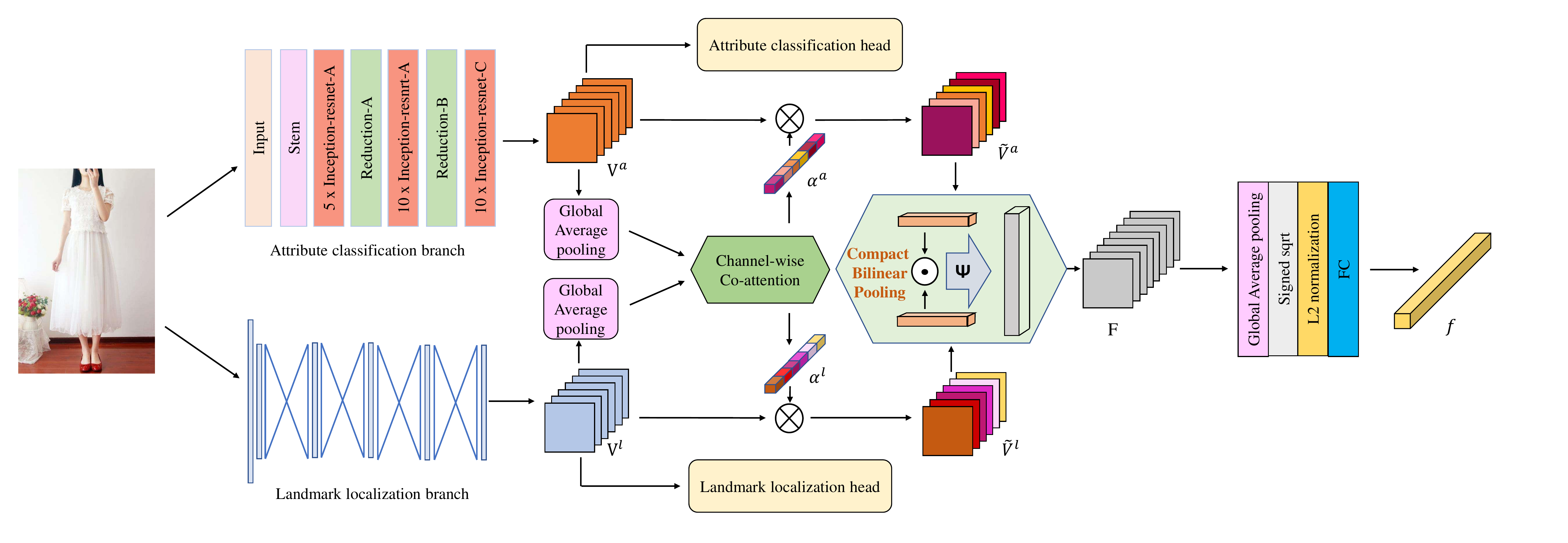}
%\vspace{-3em}
% \setlength{\abovecaptionskip}{-10pt}
% \setlength{\belowcaptionskip}{-5pt}
\caption{Overview of our proposed Attentional Heterogeneous Bilinear Network (AHBN).
The input image is simultaneously fed into two heterogeneous branches, \ie, an attribute classification network and a landmark localization network, to extract both attribute descriptors and part location information.
These two branches are individually driven by a multi-label attribute classification task and an eight-landmark detection task to ensure their specific purposes.   
To focus on mutually correlated channels, 
a channel-wise co-attention module 
is proposed to take global feature representations of the two branches as input and 
output the channel weights for both branches.
After multiplying the weights on feature channels, the resulting dual-branch feature maps
are further integrated via compact bilinear pooling in a spatial-wise fashion, followed by average pooling. Finally, a global feature vector is generated as the representation of the input fashion image, and trained with an ID classification loss.
 $\otimes$ represents the operation that multiplies each feature map by a weight. 
}
\label{fig:arch}
\end{figure*}

\section{Related Work}
\noindent {\bf Fashion Retrieval.} 
Fashion product retrieval based on images~\cite{huang2015cross,hadi2015buy,liu2016deepfashion,gu2018multi,liang2016clothes,li2017mining,zhang2017trip,zhao2016clothing,wang2020deep,nie2020deep,wang2019sketch,peng2019unsupervised} or videos~\cite{cheng2017video2shop,garcia2017dress} has attracted an increasing attention, along with the development of e-commerce. To further add an interaction between users and machines, the task of 
fashion search with attribute manipulation~\cite{han2017automatic,zhao2017memory,ak2018learning} allows the user to provide additional descriptions about wanted attributes that are not presented in the query image.

%\textcolor{red}{
Many excellent methods have been explored for the retrieval task. Wang et al.~\cite{wang2020deep} proposed a deep hashing method with pairwise similarity-preserving quantization constraint, termed Deep Semantic Reconstruction Hashing (DSRH), which defines a high-level semantic affinity within each data pair to learn compact binary codes. Nie et al.\cite{nie2020deep} designed different network branches for two modalities and then adopt multiscale fusion models for each branch network to fuse the multiscale semantics. Then multi-fusion models also embed the multiscale semantics into the final hash codes, making the final hash codes more representative. Wang et al.\cite{wang2019sketch} used blind feedback in an unsupervised method in order to make the re-ranking approach invisible to users and adaptive to different types of image datasets. Peng et al.\cite{peng2019unsupervised} transfered knowledge from the source domain to improve cross-media retrieval in the target domain. 
%}

Some works in~\cite{huang2015cross,corbiere2017leveraging,liu2016deepfashion,zhang2017trip,li2017mining} improve performance of fashion retrieval by incorporating additional semantic information such as attributes, categories or textual descriptions \etc. 
Some works focus on training a fashion retrieval model with specifically designed losses~\cite{song2017learning,jiang2018deepproduct,jiang2016deep,gu2018multi}. 
There are also efforts on optimizing the feature representation \cite{simo2016fashion,hsiao2017learning,song2017learning}. 
Attention mechanisms have also been employed in fashion product retrieval to focus on important image regions~\cite{ji2017cross}. 

As for fashion retrieval datasets, two public large-scale fashion datasets, DeepFashion \cite{liu2016deepfashion} and Exact Street2Shop \cite{hadi2015buy}, contribute to the development of fashion retrieval. DeepFashion \cite{liu2016deepfashion} collects over $800K$ images with rich annotated information, including attributes, landmarks and bounding boxes.
Exact Street2Shop Dataset \cite{hadi2015buy} is split into two categories: $20,357$ street photos and $404,683$ shop photos for fashion retrieval applications
\cite{hadi2015buy,wang2017clothing,xiong2016parameter,jiang2018deep,jiang2016deep,song2017learning,jiang2018deepproduct}.

Among the above mentioned approaches, FashionNet~\cite{liu2016deepfashion} is most similar to our approach, which also incorporates both attribute and landmark information for retrieval. However, our method integrates the attribute and landmark information in a more systematic way via the proposed attentional bilinear pooling module. The mutual interaction between the two information sources is not only used to jointly select important feature channels, but also employed to form a bilinear final representation.

\noindent {\bf Bilinear Pooling Networks.}
Lin et al.~\cite{lin2015bilinear} proposed a bilinear CNN model and successfully applied it to fine-grained visual recognition.
The model consists of two CNN-based feature extractors, whose outputs are further integrated by the outer product at each location and average pooling across locations. 
Differing from the element-wise product, the employed outer product is capable of modeling pairwise interactions between all elements of both input vectors.  
Note that this architecture is related to the two-stream hypothesis of human visual system~\cite{goodale1992separate}, with two pathways corresponding to identification and localization respectively.
However, the original bilinear pooling computes outer products and yields very high-dimensional representations, which makes it computationally expensive.
To this end, Gao et al.~\cite{gao2016compact} proposed Compact Bilinear Pooling (CBP) using sampling-based low-dimensional approximations of the polynomial kernel, which reduces the dimensionality by two orders of magnitude with little loss of performance.
Fukui et al.~\cite{fukui2016multimodal} extended CBP~\cite{gao2016compact} to the multimodal case, and applied their proposed Multimodal Compact Bilinear (MCB) pooling to visual question answering and visual grounding.
Kim et al.~\cite{kim2016hadamard} proposed Multimodal Low-rank Bilinear (MLB) pooling to reduce the high dimensionality of full bilinear pooling using a low-rank approximation.
Multimodal Factorized Bilinear (MFB) pooling~\cite{yu2017multi} can be considered as a generalization of MLB, which has a more powerful representation capacity with the same output dimensionality. 

Many bilinear models rely on two homogeneous branches, \eg, two similar networks, and do not explicitly assign different roles to them.
By contrast, in our design, two heterogeneous branches are adopted and their auxiliary tasks/losses ensure that they can extract information from different perspectives. {In this sense, compared to bilinear networks with homogeneous branches, 
our heterogeneous model behaves more like the two-stream visual processing system of human brain~\cite{goodale1992separate}.}

\noindent {\bf Attention Mechanism.} 
Bahdanau et al.~\cite{bahdanau2014neural} proposed to use an attention mechanism in sequence-to-sequence model, to focus on relevant parts from the input sequence adaptively at each decoding time-step. 
Xu et al.~\cite{xu2015show} introduced two attention mechanisms into image caption, namely soft attention and hard attention. 
Soft attention is differential and so it can be trained end-to-end. 
Based on the work of~\cite{xu2015show}, Luong et al.~\cite{luong2015effective} proposed global attention and local attention. Global attention simplifies the soft attention and local attention is the combination of soft and hard attention mechanisms.
Vaswani et al.~\cite{vaswani2017attention} proposed the self-attention mechanism, which computes the pairwise relevance between different parts of input. 
Lu et al.~\cite{lu2016hierarchical} proposed a co-attention module for visual question answering that jointly performs visual attention and question attention.
Different from spatial attention that selects image sub-regions, the channel-wise attention mechanism~\cite{chen2017sca} computes weights for convolutional feature channels and can be viewed as a process of selecting CNN filters or semantic patterns.
The Squeeze-and-Excitation (SE) block~\cite{hu2018squeeze} can be also considered as a case of channel-wise attention, where a global image representation is used to guide the generation of channel weights.

Note that the SE block~\cite{hu2018squeeze} is self-guided, as it is used for single-branch architectures.
In contrast, our proposed channel-wise co-attention mechanism first constructs a joint representation of two branches, 
and uses it to guide the channel weights generation for both branches.
In other words, the two branches are mutually guided in our proposed co-attention module.

\section{Model Architecture}
In this section, we give a detailed introduction of our proposed Attentional Heterogeneous Bilinear Network (AHBN) for fashion retrieval, including the overall structure and its three main components (\ie, an attribute classification branch, a landmark localization branch and an attentional bilinear network).

\subsection{Overall Structure}
As shown in Figure~\ref{fig:arch}, 
the input image is firstly fed into a two-branch architecture: an attribute classification branch to extract attribute visual descriptions and a landmark localization branch to detect part locations.
The resulting two feature maps ${\mathbf{V}^{l}}$ and ${\mathbf{V}^{a}}$ are rescaled to the same spatial size (\eg, $8 \times 8$) via average pooling.
Note that ${\mathbf{V}^{l}}$ and ${\mathbf{V}^{a}}$ are the activation before the final classification/localization layers.
A channel-wise co-attention mechanism is then applied to adaptively and softly select feature channels of ${\mathbf{V}^{l}}$ and ${\mathbf{V}^{a}}$, 
where the guidance signal is a joint representation of both feature maps.
The pairwise interactions between all channels of the weighted feature maps
are modeled by CBP~\cite{gao2016compact} at each location.
The final global representation of the input image is then 
obtained by applying average pooling across all locations of the CBP~\cite{gao2016compact} output.
An ID classification loss is used to supervise the final representation.

During training, the two branches are firstly pre-trained with their respective auxiliary losses, 
and then the whole model is end-to-end trained with both the final and auxiliary losses.
At test time, the similarity between two images is calculated based on the Euclidean distance of their final representations.

\subsection{Attribute Classification Branch} \label{subsection:attribute}

The attribute classification branch is based on the Inception-ResNet-v2 network~\cite{szegedy2017inception} which is a combination of the Inception architecture~\cite{rethink16cvpr} and residual connections~\cite{He2016Deep}.
To be specific, the filter concatenation module in the original Inception architecture is replaced by residual connections.
This hybrid network not only leads to improved recognition performance, but also achieves faster training speed.
%Because of residual connection, the coverage rate of the Inception network is greatly accelerated  and a slight increase in performance is obtained.

We adopt the binary cross entropy (BCE) loss for the multi-label attribute classification, which is defined as follows:
\begin{align}
\label{eq:bce_loss}
L_{\mbox{attribute}} &= \frac{1}{N}\sum_{i=1}^{N}l_{i}, \\
\mbox{where} \quad l_{i} &= -\big( y_{i} \cdot\mathrm{log} (x_{i})+(1-y_{i})\cdot\mathrm{log} (1-x_{i})\big) \notag,
\end{align}
$N$ is the number of the attributes. $l_{i}$ is the BCE loss for the $i$-th attribute. $y_{i}$ $\in$ $\{{0,1}\}$ and $x_{i}$ $\in$ $\big({0,1}\big)$ are the ground truth and the prediction score for the $i$-th attribute respectively.

\subsection{Landmark Localization Branch} \label{subsection:landmark}

%\textcolor{red}{
Recently, many novel localization methods have been proposed~\cite{hong2018multimodal,yu2019hierarchical,hong2015multimodal}. Hong et al.~\cite{hong2018multimodal} proposed a novel face pose estimation method based on feature extraction with improved deep neural networks and multi-modal mapping relationship with multi-task learning. Different modals of face representations are naturally combined to learn the mapping function from face images to poses. Yu et al.~\cite{yu2019hierarchical} integrated sparse constraints and an improved RELU operator to address click feature prediction from visual features. Hong et al.~\cite{hong2015multimodal} proposed a pose recovery method, i.e., non-linear mapping with multi-layered deep neural network for video-based human pose recovery. It is based on feature extraction with multi-modal fusion and back-propagation deep learning.
%}

As with most existing landmark localization approaches, we also transform the task into the heatmap regression problem. In this paper, our landmark localization network is based on stacked hourglass architecture~\cite{newell2016stacked}, which consists of a $7\times 7$ convolution and four hourglass blocks. 
% The input image size of this network is set to $256 \times 256$. 
The last feature map before generating heatmaps is of size $256 \times 64 \times 64$.  

The hourglass network~\cite{newell2016stacked} can obtain the information of all scale images. It is named because
the down sampling and up sampling of the network look like an hourglass from the structure. The design of the structure is mainly derived from the need to grasp the information of each scale. Hourglass is a simple, minimal design with the ability to capture all feature information and make final pixel level predictions.
%The size of the output heatmap is $ {8\times 64\times 64}$ for the DeepFashion dataset with $8$ landmarks.

Considering that the visibility of each landmark for each input is different, we designed our loss function as follows:
\begin{equation}
\label{eq:lm_loss}
 L_{\mbox{landmark}} = \sum_{m=1}^{M}v_{m}\big \| X_{m}-Y_{m} \big \|,
\end{equation}
where $M$ means the number of annotated landmarks, $\big \| \cdot  \big \|$ represents the Euclidean distance. $v_{m}\in \left \{0, 1\right \}$, $X_{m}\in \mathbb{R}^{64\times 64}$, $Y_{m}\in \mathbb{R}^{64\times 64}$ represent respectively the visibility of the $m$-th landmark, the predicted heatmap and the ground-truth heatmap. For the DeepFashion dataset, $M = 8$.

\begin{figure}
\centering
\includegraphics[width=0.85\linewidth]{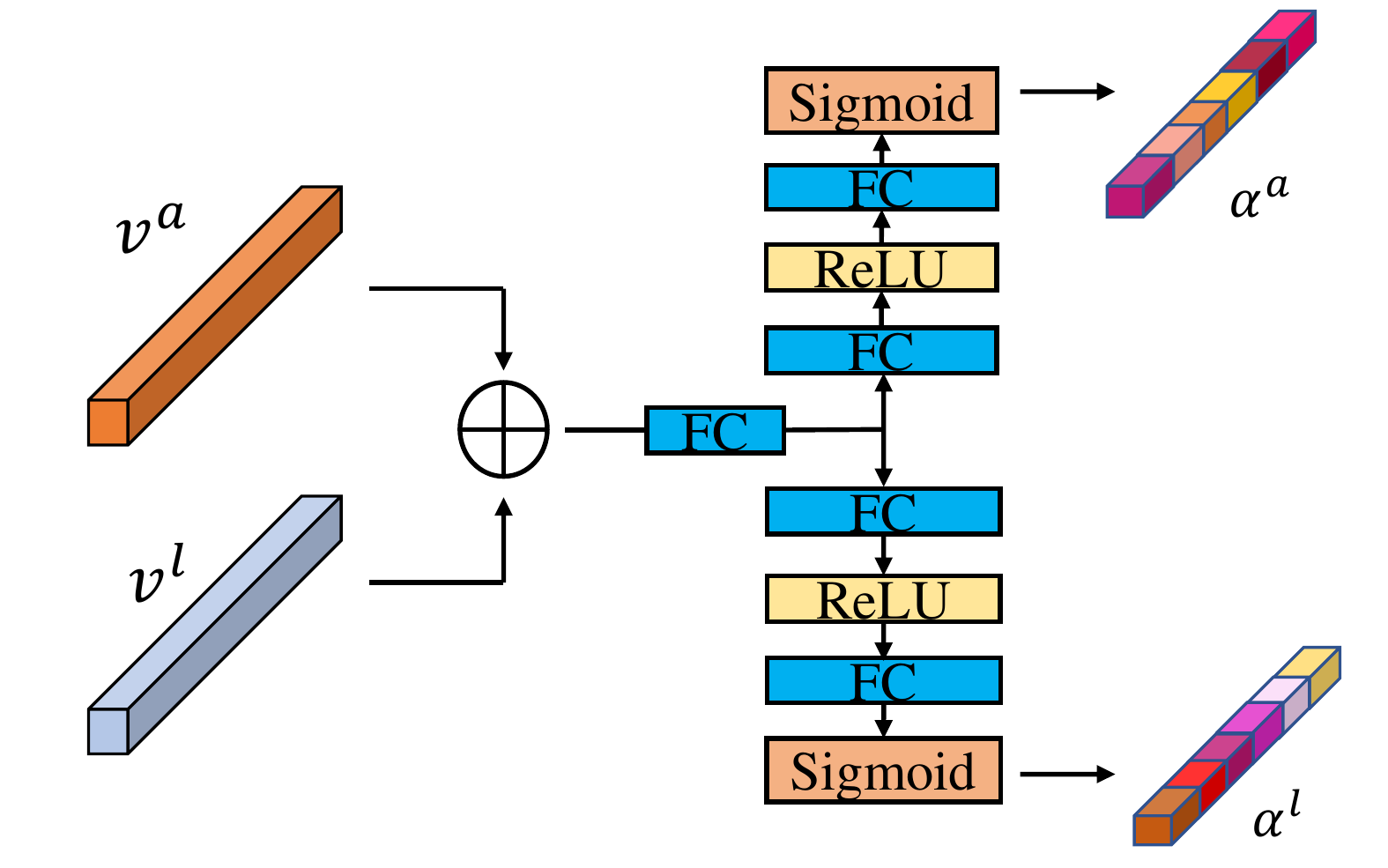}
%\vspace{-1em}
\caption{Our Proposed Channel-Wise Co-Attention Module. It takes global representations of the two branches ($\mathbf{v}^a$ and $\mathbf{v}^l$) as inputs,
passes them through shared and non-shared layers, and generates channel weights for both branches ($\alpha^a$ and $\alpha^l$). 
$\oplus$ indicates the concatenation operation.
}
\label{fig:chan}
\end{figure}

% Algorithm
%\resizebox{0.86\textwidth}{!}{
\begin{algorithm}
%\footnotesize
\caption{Attentional Bilinear Network}
\label{alg:ahbn}
\begin{algorithmic}[1]
    \Require ${\mathbf{V}^{a}}\in\mathbb{R}^{C^{a} \times W^{a} \times H^{a}}, {\mathbf{V}^{l}}\in\mathbb{R}^{C^{l} \times W^{l} \times H^{l}}$
    \Ensure $F\in\mathbb{R}^{d \times W \times H}$
    \Function {$AHBN$}{${\mathbf{V}^{a}}, {\mathbf{V}^{l}}, W, H, d$}
    \State ${\widetilde{\mathbf{V}}^{a}}, {\widetilde{\mathbf{V}}^{l}} = CoATTEN({\mathbf{V}^{a}}, {\mathbf{V}^{l}})$
    \State Re-sample ${\widetilde{\mathbf{V}}^{a}}$, ${\widetilde{\mathbf{V}}^{l}}$ to the same spatial size $(W \times H)$
    %\State average pool ${\widetilde{\mathbf{V}}^{a}}$ to $C^{a} \times W \times H$
    %\State average pool ${\widetilde{\mathbf{V}}^{l}}$ to $C^{l} \times W \times H$
    \For{$i = 1, ...,W$}
        \For{$j = 1, ...,H$}
            \State // Define ${x_{ij}^{a}}$ as the local feature at the $(i,j)$-th location of ${\widetilde{\mathbf{V}}^{a}}$.
            \State // Define ${x_{ij}^{l}}$ as the local feature at the $(i,j)$-th location of ${\widetilde{\mathbf{V}}^{l}}$.
            \State $F_{ij} = CBP({x_{ij}^{a}}, {x_{ij}^{l}}, d)$
        \EndFor
    \EndFor
    \State \Return{$F$}
    \EndFunction
    \State
    \Function {$CoATTEN$}{${\mathbf{V}^{a}}, {\mathbf{V}^{l}}$}
    %\State average pool ${\mathbf{V}^{a}}$ to a vector ${\mathbf{v}^{a}}$
    %\State average pool ${\mathbf{V}^{l}}$ to a vector ${\mathbf{v}^{l}}$
    \State ${\mathbf{v}^{a}}=\mathrm{GlobalAveragePooling}({\mathbf{V}^{a}})$
    \State ${\mathbf{v}^{l}}=\mathrm{GlobalAveragePooling}({\mathbf{V}^{l}})$
    \State $\alpha^{a} = \mathrm{Sigmoid}({\mathbf{W}_{2}^{a}}\cdot\mathrm{Relu}({\mathbf{W}_{1}^{a}}({\mathbf{v}^{a}}\oplus{\mathbf{v}^{l}})))$
    \State $\alpha^{l} = \mathrm{Sigmoid}({\mathbf{W}_{2}^{l}}\cdot\mathrm{Relu}({\mathbf{W}_{1}^{l}}({\mathbf{v}^{a}}\oplus{\mathbf{v}^{l}})))$
    \State ${\widetilde{\mathbf{V}}^{a}} = {\mathbf{V}^{a}}\otimes\alpha^{a}$
    \State ${\widetilde{\mathbf{V}}^{l}} = {\mathbf{V}^{l}}\otimes\alpha^{l}$
    \State \Return{${\widetilde{\mathbf{V}}^{a}}$ and ${\widetilde{\mathbf{V}}^{l}}$}
    \EndFunction
    \State
    \Function {$CBP$}{${x_{1}}, {x_{2}}, d$}
    \State $y_{1} = Project({x_{1}},d)$
    \State $y_{2} = Project({x_{2}},d)$
    \State $F =  FFT^{-1}(FFT(y_{1}) \circ FFT(y_{2}))$
    \State \Return{$F$}
    \EndFunction
    \State
    \Function {$Project$}{$x, d$}
    \State $C = \mathrm{Length}(x)$
    \For{$k = 1 \to C$}
        \State initialize $s[k]$ from $\{+1, -1\}$ uniformly.
        \State initialize $p[k]$ from $\{1, ...,d\}$ uniformly.
    \EndFor
    \State $y = \Psi(x, s, p, d)$
    \State \Return{$y$}
    \EndFunction
    \State
    \Function {$\Psi$}{$x, s, p, d$}
    \State initialize $y$ to $[0, ...,0]^{d}$
    \For{$i=1 \to d$}
      \State $y[i] = \sum\nolimits_{t}s[t]x[t] \quad s.t. \quad p[t]=i$
    \EndFor
    \State \Return{$y$}
    \EndFunction
\end{algorithmic}
\end{algorithm}
%}

\subsection{Attentional Bilinear Network}
As shown in Sections \ref{subsection:attribute} and \ref{subsection:landmark}, 
we obtain two heterogeneous feature maps respectively driven by an attribute classification task and a landmark localization task.
In this section, we incorporate their mutual interactions to perform channel-wise attentions and generate final global representations.

The main reason of using bilinear pooling is to capture the second-order interactions between each pair of output channels from the two heterogeneous branches of our framework.
Thus, the resulting bilinear vector does not only encode the salient appearance features but also their locations.
Comparing with fully connected layer, the bilinear pooling is more effective for encoding such second-order interactions and incur much less parameters.
As the original bilinear pooling results in a long feature vector, we adopt
compact bilinear pooling (CBP)~\cite{gao2016compact} to reduce the dimension of bilinear vectors.

\noindent{\bf Channel-Wise Co-Attention.} 
Note that the feature channels of the two-branch features ${\mathbf{V}^{l}}$ and ${\mathbf{V}^{a}}$
are not equally important for a particular image.
Furthermore, the importance of a channel does not only 
depend on features in the same branch, but also is relevant to the other branch.
To this end, we propose a channel-wise co-attention mechanism as shown in Figure~\ref{fig:chan}, 
which takes global representations of two branches as inputs, models their mutual interactions, 
and outputs channel weights for both branches.
%\textcolor{red}{
To be more specific, the co-attention module takes the global representations of two branches as inputs, feeds them into a fully connected layer to encode the interaction of the two branches, and finally outputs the channel attention weights for both branches. Such that, the attention weights of each branch are determined by the two branches. In other words, the two branches mutually affect each other.
%}
It is shown in our experiments that our mutually-guided co-attention module performs better than two separated self-guided attention modules.

The size of feature maps ${\mathbf{V}^{a}}$ and ${\mathbf{V}^{l}}$ are ${\mathbf{V}^{a}} \in \mathbb{R}^{C^a \times W^a \times H^a}$ 
and ${\mathbf{V}^{l}} \in \mathbb{R}^{C^l \times W^l \times H^l}$ 
(in our particular case, 
${\mathbf{V}^{a}}$ and ${\mathbf{V}^{l}}$ are of sizes $1536 \times 8 \times 8$ and $256 \times 64 \times 64$ respectively) are obtained by global average pooling:
\begin{align}
    {\mathbf{v}^{a}}&=\mathrm{GlobalAveragePooling}\big({\mathbf{V}^{a}}\big), \\
    {\mathbf{v}^{l}}&=\mathrm{GlobalAveragePooling}\big({\mathbf{V}^{l}}\big),
\end{align}
where ${\mathbf{v}^{a}} \in \mathbb{R}^{C^a}$ and ${\mathbf{v}^{l}} \in \mathbb{R}^{C^l}$.
These two representations are concatenated and fed into two Multi-Layer Perceptions to calculate channel-wise attention weights for two branches:
\begin{align}
    \alpha^{a}&=\mathrm{Sigmoid}\big({\mathbf{W}_{2}^{a}}\cdot\mathrm{Relu}\big({\mathbf{W}_{1}^{a}}\big({\mathbf{v}^{a}}\oplus{\mathbf{v}^{l}}\big)\big)\big),\\
    \alpha^{l}&=\mathrm{Sigmoid}\big({\mathbf{W}_{2}^{l}}\cdot\mathrm{Relu}\big({\mathbf{W}_{1}^{l}}\big({\mathbf{v}^{a}}\oplus{\mathbf{v}^{l}}\big)\big)\big),
\end{align}
where ${\mathbf{W}_{1}^{a}}\in\mathbb{R}^{k^{a} \times C}$, ${\mathbf{W}_{1}^{l}}\in\mathbb{R}^{k^{l} \times C}$, ${\mathbf{W}_{2}^{a}}\in\mathbb{R}^{C^{a} \times k^{a}}$ and ${\mathbf{W}_{2}^{l}}\in\mathbb{R}^{C^{l} \times k^{l}}$ are linear transformation matrices (biases in linear transformations are omitted here). $k^{a}$ and $k^{l}$ are the projection dimensions. $\oplus$ denotes the concatenation operation and $C=C^{a}+C^{l}$. 
$\alpha^{a}$ and $\alpha^{l}$ are the channel-wise attention weights for the attribute classification branch and the landmark localization branch respectively.
Besides Sigmoid, we also experiment with Softmax to compute the weights, which, however yields worse performance.
The reason may be that the importance of different feature channels is not mutually exclusive.

Finally, we obtain two weighted feature maps as follows:
\begin{align}
%\begin{split}
    {\widetilde{\mathbf{V}}^{a}} &= {\mathbf{V}^{a}}\otimes\alpha^{a}, \\
    {\widetilde{\mathbf{V}}^{l}} &= {\mathbf{V}^{l}}\otimes\alpha^{l},
%\end{split}
\end{align}
where $\otimes$ represents the operation that multiplies each feature map by its corresponding channel weight. 
Before processed by the following spatial-wise compact bilinear pooling layer, $\mathbf{V}^{a}$ and $\mathbf{V}^{l}$
are re-sampled to the same spatial size ($W \times H$). In our case, $W = H = 8$.

\noindent{\bf Spatial-Wise Compact Bilinear Pooling.}
At each of the $W \times H$ spatial locations, 
we now have a vector encoding visual attribute information (\ie, ``how to describe'') 
and a vector representing object-part location information (\ie, ``where to look'').
In this section, we adopt Compact Bilinear Pooling with count sketch to 
model their multiplicative interactions between all elements of the two vectors.

Given a local feature vector $x_{ij}\in\mathbb{R}^{k}$ at the $(i,j)$-th location of the feature map, the count sketch function $\Psi$ \cite{charikar2002finding} projects $x_{ij}$ to a destination vector $y_{ij}\in\mathbb{R}^{d}$.
%value of which is initialized to zero. 
Moreover, a signed vector $s\in\mathbb{Z}^{k}$ and a mapping vector $p\in\mathbb{N}^{k}$ are employed in the sketch function. The value of $s$ is randomly selected from $\{+1,-1\}$ by equal probability and $p$ is randomly sampled from $\{1,...,d\}$ in a uniformly distributed way. Then the $\Psi$ can be defined as follows:
\begin{equation}
\begin{split}
    y_{ij} &= \Psi(x_{ij}, s, p) = [v_{1},...,v_{d}], \\
    \mbox{where} \quad v_{t} &= \sum\nolimits_{l}s[l] \cdot x_{ij}[l] \quad s.t. \quad p[l]=t.
\end{split}
\end{equation}
The count sketch function taking the outer product of two vectors $x_{ij}^{a}$ and $x_{ij}^{l}$ as input can be written as the convolution of count sketches of individual vectors:
\begin{equation}
    \Psi(x_{ij}^{a} \odot x^{l}_{ij}, s, p) = \Psi(x_{ij}^{a}, s, p) \ast \Psi(x_{ij}^{l}, s, p),
\end{equation}
where $\odot$ represents the outer product operation and $\ast$ refers to the convolution operation.
where $\ast$ refers to the convolution operation.
Finally, we can get the bilinear feature by transforming between time domain and frequency domain:

\begin{equation}
\setlength{\abovedisplayskip}{-6pt}
\setlength{\belowdisplayskip}{3pt}
\begin{split}
    F_{ij} =& FFT^{-1}\Big(FFT(\Psi(x_{ij}^{a}, s_{ij}^{a}, p_{ij}^{a})) \circ \\
    &FFT(\Psi(x_{ij}^{l}, s_{ij}^{l}, p_{ij}^{l}))\Big), 
\end{split}
\end{equation}

where $\circ$ represents element-wise multiplication. The overall algorithm of our proposed attentional bilinear network is shown in Algorithm~\ref{alg:ahbn}.
% The overall algorithm of our proposed attentional bilinear network is shown in Algorithm~\ref{alg:ahbn}.

\noindent{\bf ID Classification and Optimization.}
The resulting feature map $F$ is then transformed to a global image representation $f$, 
using a series of operations consisting of global average pooling, signed square root, $l_2$-norm normalization and a fully connected layer.

The final image representation is then employed to perform an ID classification task, which considers each clothes instance as a distinct class. 
To do so, we further add a linear transformation layer to project the global representation to a vector whose dimension equals to the number of ID classes. 
The cross-entropy loss is employed as follows:
\begin{equation}
\label{eq:ce_loss}
%\begin{split}
    % \mathrm{Softmax}(x_{i}) = \mathrm{log}\frac{\exp(x_{i}){\sum_j\exp(x_{j})} \\
    L_{\mbox{ce}}(x, gt) = -\log\left(\frac{\exp(x[gt])}{\sum_i \exp(x[i])}\right),
%\end{split}
\end{equation}
where $x$ is the prediction vector and $gt$ is the index of the ground truth class. Note that the whole framework can be end-to-end trained only with this ID classification task.
But in practice, we train our full AHBN model with all the losses, including \eqref{eq:bce_loss}, \eqref{eq:lm_loss} and \eqref{eq:ce_loss}, to ensure that the two branches achieve their respective tasks.  
At test time, we only compute the $2048$D global representations of query and gallery images, and the corresponding Euclidean distance.

\section{Experiments}
In this section, we validate the effectiveness of our proposed method on two public datasets for fashion product retrieval, 
\ie, DeepFashion \cite{liu2016deepfashion} and Exact Street2Shop \cite{hadi2015buy}. 
An ablation study is conducted to investigate the contributions of individual components in our proposed architecture.
Our approach also outperforms other evaluated methods in the three benchmarks.

\begin{figure*}
\centering
\includegraphics[width=\linewidth]{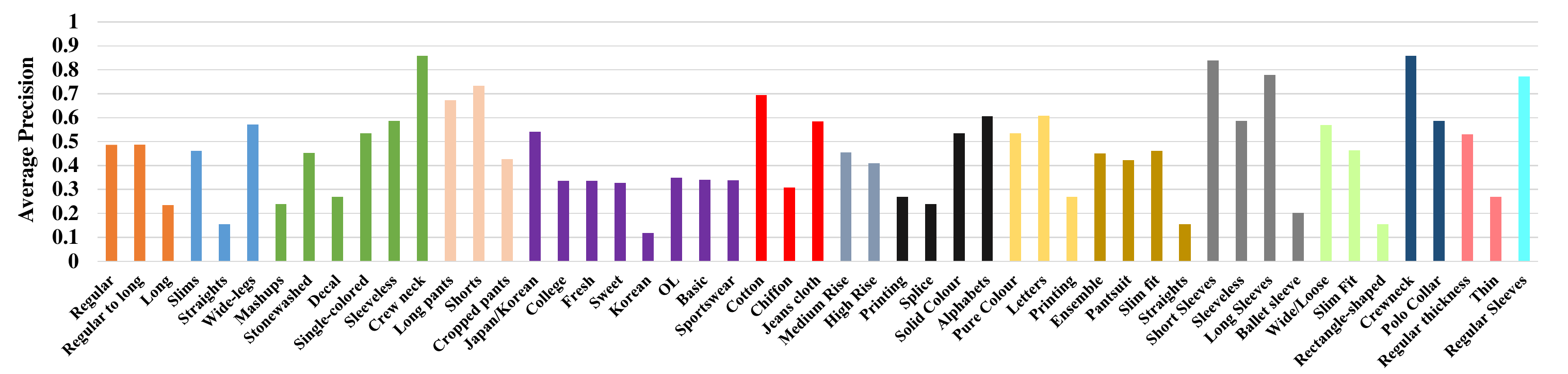}
%\vspace{-2.5em}
% \setlength{\abovecaptionskip}{-10pt}
% \setlength{\belowcaptionskip}{-5pt}
\caption{
The performance of the attribute classification branch on the Consumer-to-Shop Clothes Retrieval Benchmark in the DeepFashion Dataset.
We calculate the average precision (AP) for each attribute. The figure shows both the attribute label and its corresponding AP. 
And different colors of bars indicate different attribute types. 
From left to right, the corresponding attribute types are length of upper-body clothes, trousers, part details, length of trousers, style, fabric, waistlines, texture, graphic elements, length of sleeves, design of dresses, fitness, collars, thickness, and sleeves.
The mean average precision of attributes classification is $46.0\%$.
}
\label{fig:attribute}
\end{figure*}

\begin{figure}[t]
\centering
\includegraphics[width=0.95\linewidth]{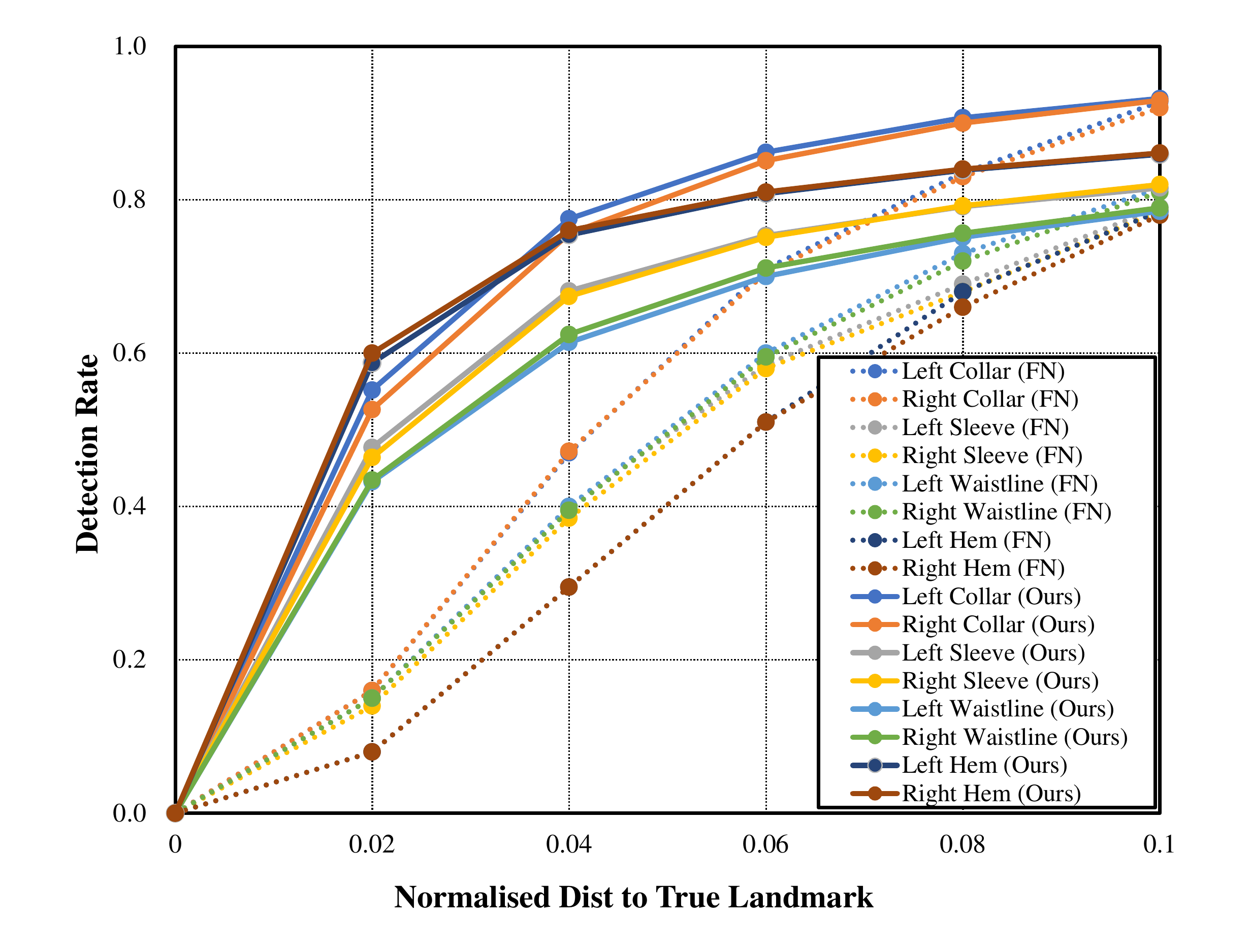}
%\vspace{-1em}
% \setlength{\abovecaptionskip}{-5pt}
% \setlength{\belowcaptionskip}{-5pt}
\caption{The performance of the landmark localization branch on DeepFashion Consumer-to-Shop Benchmark. We calculate the normalized mean error (NME) for each of 8 landmarks. In the figure, the dotted line and the solid line are the results of FashionNet \cite{liu2016deepfashion} and our stacked hourglass network respectively, and the same landmark is represented by the same color for both lines. The performance of our method significantly outperforms FashionNet on NME for each of 8 landmarks.}
\label{fig:landmark}
\end{figure}

\subsection{Datasets}
The details of our adopted two large-scale datasets are described as follows.

\noindent{\bf DeepFashion.} We evaluate our model on two benchmarks in the DeepFashion dataset, \ie, the Consumer-to-Shop Clothes Retrieval Benchmark and the In-Shop Clothes Retrieval Benchmark. 
The  Consumer-to-Shop benchmark has $239,557$ cross-domain clothes images and the In-Shop benchmark has $52,712$ shop images.
Both of them have elaborated with annotated information of bounding boxes, landmarks and attributes. 
We construct the train, validation and test set in accordance with their original partition file respectively. 
For both benchmarks, we crop the region of interest for each image based on the annotated bounding boxes. 
%Sufficient attribute and landmark location information in both datasets makes it possible for our model building.

For the Consumer-to-Shop benchmark, all images are sorted into $195,540$ cross-domain pairs. 
%Besides, we re-arrange items belonging into $22723$ new items with $120968$ images for training thoughtfully and make it more difficult to pick out exact ones.
The validation set has $48,527$ images and the test set has $47,434$ images. The gallery set contains $22,669$ shop images. There are $303$ annotated attributes and the $51$ most common ones are selected for attribute classification. 

For the In-Shop benchmark, $3997$ items with $25,882$ images are for training and $3985$ items with $28,760$ images are for test. The test set contains $14,218$ query images and $12,612$ gallery images. $463$ attributes are annotated and we select the $40$ most frequent ones.

% \begin{figure}[t]
% \centering
% \includegraphics[width=0.6\linewidth]{landmark_result}
% %\vspace{-1em}
% % \setlength{\abovecaptionskip}{-5pt}
% % \setlength{\belowcaptionskip}{-5pt}
% \caption{The performance of the landmark localization branch on DeepFashion Consumer-to-Shop Benchmark. We calculate the normalized mean error (NME) for each of 8 landmarks. In the figure, the dotted line and the solid line are the results of FashionNet \cite{liu2016deepfashion} and our stacked hourglass network respectively, and the same landmark is represented by the same color for both lines. The performance of our method significantly outperforms FashionNet on NME for each of 8 landmarks.}
% \label{fig:landmark}
% \end{figure}

\noindent{\bf Exact Street2Shop.} This dataset contains street photos and shop photos of fashion products. It provides street-to-shop pairs and the clothes bounding box information for each street photo.
A detector is trained utilizing the street photos and the corresponding bounding boxes, 
and used to crop clothes bounding boxes from shop photos. 
Because this dataset does not provide landmark and attribute annotations, the attribute classification and the landmark localization branches are pretrained on the Consumer-to-Shop benchmark.
We select six clothes categories that overlap with Consumer-to-Shop to evaluate, including
dresses, leggings, outwear, pants, skirts and tops.
The partition of the training and test set is based on the original setting. 
There are $7102$ items for training and testing and $256,698$ images in the gallery.

\subsection{Implementation Details}
Our proposed model is implemented in Pytorch.
All  experiments are performed on GEFORCE GTX1080 Ti graphics processing units. 
%Iterations for training is about $210k$ before our model converges. 
%The size of image feature vector sets as $2048$ dimension uniformly.
The dimensionality of the final global representation is set to $2048$.
We first pre-train the attribute classification branch with loss \eqref{eq:bce_loss} 
and the landmark localization branch with loss \eqref{eq:lm_loss}, 
and then train the full AHBN model with three loss functions \eqref{eq:bce_loss}, \eqref{eq:lm_loss} and \eqref{eq:ce_loss}. 
We use Adam as the optimizer.
The batch size is set to $20$ and the maximum epoch number is $35$. 
The learning rate is initialized to $0.0001$ and reduced by half after every $5$ epochs. 
Data augmentation is adopted during training, such as horizontal flip and random rotation.

Following \cite{liu2016deepfashion,hadi2015buy}, we calculate top-$k$ accuracies for every query image.
Given a query image, we calculate Euclidean distances between it and all images in the gallery set.
Then, we obtain top-$k$ results by ranking these distances in an ascending order and the retrieval will be considered as a success if the ground-truth gallery image is found in the top-$k$ results.

\begin{table*}[t]
\renewcommand\arraystretch{0.9}
\centering
    \caption{Ablation study on the DeepFashion Consumer-to-Shop Benchmark. The ``ATTR'' and ``LM'' columns indicate if an algorithm uses the corresponding attribute and landmark annotation.}
    \label{top:accuracies}
    \resizebox{0.95\textwidth}{!}{
    \begin{tabular}{lccccccccc}
    \toprule
    Model&ATTR&LM&Attention&Acc@1&Acc@10&Acc@20&Acc@30&Acc@40&Acc@50\\
    \midrule
    Single-Branch & $\times$ & $\times$ & $\times$ & $0.239$ & $0.498$ & $0.564$ & $0.602$ & $0.628$ & $0.648$\\
    Single-Branch + Res50 & $\times$ & $\times$ & $\times$ & $0.142$ & $0.367$ & $0.449$ & $0.501$ & $0.538$ & $0.568$\\
    Single-Branch + Spatial Atten. & $\times$ & $\times$ & $\times$ & ${0.232}$ & ${0.475}$ & ${0.540}$ & ${0.576}$ & ${0.601}$ & ${0.624}$\\
    Single-Branch + ATTR & $\checkmark$ & $\times$ & $\times$ & $0.243$ & $0.504$ & $0.578$ & $0.609$ & $0.634$ & $0.654$\\
    Single-Branch + LM & $\times$ & $\checkmark$ & $\times$ & $0.118$ & $0.297$ & $0.357$ & $0.396$ & $0.425$ & $0.447$\\
    Two-Branch w. $8$LM & $\checkmark$ & $\checkmark$ & $\times$ & $0.241$ & $0.501$ & $0.566$ & $0.604$ & $0.631$ & $0.651$\\
    Two-Branch w. $256$LM & $\checkmark$ & $\checkmark$ & $\times$ & $0.256$ & $0.524$ & $0.591$ & $0.628$& $0.655$ & $0.674$\\
    Two-Branch w. $256$LM + BP & $\checkmark$ & $\checkmark$ & $\times$ & ${0.247}$ & ${0.502}$ & ${0.568}$ & ${0.602}$ & ${0.631}$ & ${0.652}$\\
    Two-Branch w. $256$LM + Cat & $\checkmark$ & $\checkmark$ & $\times$ & $0.247$ & $0.516$ & $0.582$ & $0.619$ & $0.644$ & $0.663$\\
    Two-Branch w. $256$LM + Mul & $\checkmark$ & $\checkmark$ & $\times$ & $0.244$ & $0.514$ & $0.580$ & $0.618$ & $0.645$ & $0.664$\\
    Two-Branch w. $256$LM + Sum & $\checkmark$ & $\checkmark$ & $\times$ & $0.243$ & $0.510$ & $0.573$ & $0.610$ & $0.635$ & $0.653$\\
    Two-Branch w. $256$LM + Sepa. Atten. & $\checkmark$ & $\checkmark$ & Separate Atten. & $0.257$ & $0.528$ & $0.595$ & 0$.633$ & $0.660$ & $0.679$\\
    Our AHBN Model + $4\times4$ & $\checkmark$ & $\checkmark$ & Co-Attention & ${0.250}$ & ${0.507}$ & ${0.572}$ & ${0.606}$ & ${0.634}$ & ${0.655}$\\
    Our AHBN Model + $16\times16$ & $\checkmark$ & $\checkmark$ & Co-Attention & ${0.248}$ & ${0.505}$ & ${0.571}$ & ${0.605}$ & ${0.633}$ & ${0.654}$\\
    Our AHBN Model + Res50 & $\checkmark$ & $\checkmark$ & Co-Attention & $0.214$ & $0.461$ & $0.529$ & $0.567$ & $0.594$ & $0.615$\\
    Our AHBN Model + Softmax & $\checkmark$ & $\checkmark$ & Co-Attention & ${0.242}$ & ${0.495}$ & ${0.560}$ & ${0.596}$ & ${0.623}$ & ${0.643}$\\
    Our AHBN Model & $\checkmark$ & $\checkmark$ & Co-Attention & $\mathbf{0.260}$ & $\mathbf{0.535}$ & $\mathbf{0.603}$ & $\mathbf{0.640}$ & $\mathbf{0.666}$ & $\mathbf{0.686}$\\
    \bottomrule
    \end{tabular}
    \label{tab:ablation}}
\end{table*}

\subsection{Preliminary Training}
\noindent{\bf Attribute Classification Branch.} The input image size of this network is set to $299 \times 299$. And the output feature map is of size $ {1536\times 8\times 8}$.

Our attribute classification network is trained on 
the Consumer-to-Shop and In-Shop Clothes Retrieval Benchmarks. 
However, the distributions of these attributes in both datasets are extremely unbalanced.
Taking the Consumer-to-Shop Benchmark as example,
the most frequent attribute corresponds to $59,068$ images while the least frequent one is only contained in $15$ images. 
We only select top-$51$ attributes in the Consumer-to-Shop Benchmark and top-$40$ attributes in the In-Shop Benchmark respectively.

The result on the test dataset of the  Consumer-to-Shop Clothes Retrieval Benchmarks is shown in Figure~\ref{fig:attribute}. The mAP of our method is $46.0\%$. We also evaluate a Resnet50-based baseline and obtain $45.6\%$ mAP that is slightly worse than our Inception-Resnet-v2~\cite{szegedy2017inception} based model.

\noindent{\bf Landmark Localization Branch.} The input image size of this network is set to $256 \times 256$. The result on the test dataset of the  Consumer-to-Shop Clothes Retrieval Benchmarks is shown in Figure~\ref{fig:landmark}. As can be seen from Figure~\ref{fig:landmark}, the performance of our stacked hourglass network significantly outperforms FashionNet on the NME for each landmark.

\subsection{Ablation Study}
Through the ablation study in this section, 
we show the contributions of different components in our model to the final performance improvement.  
Except for our full {\bf AHBN} model, the following intermediate architectures are trained and evaluated on the Consumer-to-Shop Clothes Retrieval Benchmark in DeepFashion.

\noindent{\bf Single-Branch} The single-branch architecture (Inception-ResNet-v2~\cite{szegedy2017inception}) trained only with the ID classification loss.   
The whole network is pre-trained on ImageNet except for the last linear transformation layer.
For fair comparison, the final image representation is set to have a dimension of $2048$.

\noindent{\bf Single-Branch + Res50} The single-branch architecture (Resnet50) trained only with the ID classification loss.   
The whole network is pre-trained on ImageNet except for the last linear transformation layer.

\noindent{\bf Single-Branch + Spatial Atten.} The single-branch architecture (Inception-ResNet-v2~\cite{szegedy2017inception}) trained with spatial attention. 
The whole network is pre-trained on ImageNet except for the last linear transformation layer.

\noindent{\bf Single-Branch + ATTR} The single-branch architecture (Inception-ResNet-v2~\cite{szegedy2017inception}) trained with the ID classification and the multi-label attribute classification jointly.

\noindent{\bf Single-Branch + LM} The single-branch architecture (hourglass) trained with the ID classification and the landmark localization jointly.

\noindent{\bf Two-Branch w. $8$LM} The two-branch model with $8$-channel landmark heatmaps. 
In this model, we adopt the final layer of the landmark localization branch, which corresponds to the $8$ explicit landmarks to be predicted as $\mathbf{V}^l$.
CBP~\cite{gao2016compact} is employed in this model to integrate the two-branch features, but the channel-wise co-attention mechanism is disabled.

\noindent{\bf Two-Branch w. $256$LM} The two-branch architecture with $256$-channel landmark feature maps. 
Instead of using the heatmap for the explicit $8$ landmarks, we employ the $256$-dimensional feature maps just before the final prediction of the landmark branch. The channel-wise co-attention mechanism is also disabled in this model.

\noindent{\bf Two-Branch w. $256$LM + BP} The two-branch architecture with $256$-channel landmark feature maps.
The compact bilinear pooling is replaced by the standard bilinear pooling network.
Instead of using the heatmap for the explicit $8$ landmarks, we employ the $256$-dimensional feature maps just before the final prediction of the landmark branch. The channel-wise co-attention mechanism is also disabled in this model.

\noindent{\bf Two-Branch w. $256$LM + Cat} The two-branch architecture with $256$-channel landmark feature maps. Instead of CBP~\cite{gao2016compact}, concatenation is employed in this model to integrate the two-branch features.

\noindent{\bf Two-Branch w. $256$LM + Mul} The two-branch architecture with $256$-channel landmark feature maps. Instead of CBP~\cite{gao2016compact}, element-wise multiplication is employed in this model to integrate the two-branch features.

\noindent{\bf Two-Branch w. $256$LM + Sum} The two-branch architecture with $256$-channel landmark feature maps. Instead of CBP~\cite{gao2016compact}, element-wise summation is employed in this model to integrate the two-branch features.

\noindent{\bf Two-Branch w. $256$LM + Sepa. Atten.} In this model, the channel-wise co-attention mechanism is replaced by two separated self-guided channel attention modules, which are similar to two Squeeze-and-Excitation blocks~\cite{hu2018squeeze}. 

\noindent{\bf Our AHBN Model + $4\times4$} In this model, the H and W setting are both $4$ after average pooling layers.

\noindent{\bf Our AHBN Model + $16\times16$} In this model, the H and W setting are both $16$. As the output size of Inception-ResNet-v2 and Hourglass is $8\times8$ and $64\times64$ respectively, we employ a 
upsampling layer after Inception-ResNet-v2 to raise the size to $16\times16$ and a average pooling layer after Hourglass to reduce the size to$16\times16$.

\noindent{\bf Our AHBN Model + Res50} In this model, the backbone is replaced by Resnet50.

\noindent{\bf Our AHBN Model + Softmax} In this model, the Sigmoid function is replaced by the Softmax function for calculating the channel attention weights.
%\noindent{\bf AHBN} Our proposed full model as shown in Figure~\ref{fig:arch}.  

\begin{table*}[t]
\centering
  \caption{The Comparison of Top-$20$ Retrieval Accuracies on the Exact Street2Shop Dataset. The ``ATTR'' and ``LM'' columns indicate if an algorithm uses the corresponding attribute and landmark annotation.}
  \label{modelonWTBI:accuracy}
  \resizebox{0.95\textwidth}{!}{
  \begin{tabular}{ccccccccc}
    \toprule
    Model&ATTR&LM&Dresses&Leggings&Outerwear&Pants&Skirts&Tops\\
    \midrule
    WTBI \cite{hadi2015buy} & $\checkmark$ & $\times$ & $0.371$ & $0.221$ & $0.210$ & $0.292$ & $0.546$ & $0.381$\\
    Impdrop+GoogLeNet \cite{wang2017clothing} & $\times$ & $\times$ & $0.621$ & $-$ & $-$ & $-$ & $0.709$ & $0.523$\\
    Xiong et al. \cite{xiong2016parameter} & $\times$ & $\times$ & $0.583$ & $-$ & $0.509$ & $-$ & $0.736$ & $0.470$\\
    Jiang et al. \cite{jiang2018deep} & $\times$ & $\times$ & $0.212$ & $0.233$ & $0.224$ & $0.322$ & $0.103$ & $0.174$\\
    %VisNet \cite{shankar2017deep}  & / & / & 61.10\% & 32.40\% & 43.10\% & 31.80\% & 71.80\% & 62.60\%\\
    %Wang et al. \cite{wang2016matching}  & / & / & 56.90\% & 15.90\% & 20.30\% & 22.30\% & 50.80\% & 48.00\%\\
    % Jiang et al. \cite{jiang2016deep}  & $\times$ & $\times$ & $0.212$ & $0.145$ & $0.157$ & $0.322$ & $0.103$ & $0.174$\\
    %Unified Model \cite{song2017learning}  & $\times$ & $\times$ & 80.80\% & / & 52.20\% & 56.80\% & 76.00\% & 56.50\%\\
    R. Contrastive with Attribute~\cite{jiang2018deepproduct}  & $\times$ & $\times$ & $0.592$ & $0.201$ & $0.207$ & $0.213$ & $0.498$ & $0.471$\\
    GRNet \cite{Kuang_2019_ICCV} & $\times$ & $\times$ & $0.642$ & $-$ & $0.386$ & $0.485$ & $0.725$ & $0.583$\\
    \hline
    %Our Baseline & $\times$ & $\times$ & 84.88\% & 70.50\% & 82.04\% & 83.33\% & 91.04\% & 84.54\%\\
    %Our Bilinear Model & $\checkmark$ & $\checkmark$ & 86.24\% & 71.09\% &82.97\% & 81.82\% & 89.59\% & 84.22\%\\
    Our AHBN Model & $\checkmark$ & $\checkmark$ & $\mathbf{0.712}$ & $\mathbf{0.469}$ & $\mathbf{0.523}$ & $\mathbf{0.561}$ & $\mathbf{0.753}$ & $\mathbf{0.639}$\\
  \bottomrule
\end{tabular}
}
\end{table*}

\begin{table}[t]
\centering
  \caption{The Comparison of Top-$20$ Retrieval Accuracies on DeepFashion Consumer-to-Shop Benchmark. The ``ATTR'' and ``LM'' columns indicate if an algorithm uses the corresponding attribute and landmark annotation.}
  \label{model:accuracy}
  \resizebox{0.48\textwidth}{!}{
  \begin{tabular}{cccc}
    \toprule
    Model&ATTR&LM&Acc@20\\
    \midrule
    CtxYVGG \cite{ji2017cross} & $\checkmark$ & $\times$ & 0.479\\
    Liu et al. \cite{liu2016fashion} & $\times$ & $\checkmark$ & 0.510\\
    %CQ-CD(workshop) \cite{kuo2017feature} & $\times$ & $\times$ & 0.345\\
    Verma et al. \cite{verma2018diversity} & $\times$ & $\times$ & 0.253\\
    %Gajic et al.(workshop) \cite{gajic2018cross} & $\times$ & $\times$ & 0.450\\
    R. Contrastive with Attribute \cite{jiang2018deepproduct} & $\checkmark$ & $\times$ & 0.230\\
    %ImpDrop+GoogLeNet \cite{wang2017clothing} & $\times$ & $\times$ & 0.440*\\
    %Song et al.(workshop) \cite{song2017learning} & $\times$ & $\times$ & 0.392\\
    AMNet \cite{zhao2017memory} & $\checkmark$ & $\times$ & 0.338\\
    FashionNet \cite{liu2016deepfashion} & $\checkmark$ & $\checkmark$ & 0.188\\
    GRNet \cite{Kuang_2019_ICCV} & $\times$ & $\times$ & $\mathbf{0.644}$\\
    \hline
    %Our Baseline & $\times$ & $\times$ & 0.564\\
    %Our Bilinear Model & $\checkmark$ & $\checkmark$ & 0.580\\
    %Our Multi-task Bilinear Model & $\checkmark$ & $\checkmark$ & 0.584\\
    Our AHBN Model & $\checkmark$ & $\checkmark$ & 0.603\\
  \bottomrule
\end{tabular}}
\end{table}

\begin{table}[t]
\centering
  \caption{The Comparison of Top-$20$ Retrieval Accuracies on DeepFashion In-Shop Benchmark. The ``ATTR'' and ``LM'' columns indicate if an algorithm uses the corresponding attribute and landmark annotation.}
  \label{inshopmodel:accuracy}
  \resizebox{0.4\textwidth}{!}{
  \begin{tabular}{cccc}
    \toprule
    Model&ATTR&LM&Acc@20\\
    \midrule
    Studio2Shop \cite{lasserre2018studio2shop} & $\checkmark$ & $\times$ & 0.818\\
    DREML \cite{xuan2018deep} & $\times$ & $\times$ & 0.958\\
    % Zhou et al. \cite{zhou2017deep} & $\times$ & $\times$ & 0.701\\
    VAM \cite{wang2017clothing} & $\times$ & $\times$ & 0.923\\
    Weakly \cite{corbiere2017leveraging} & $\checkmark$ & $\times$ & 0.781\\
    Zhao et al. \cite{zhao2018adversarial} & $\times$ & $\times$ & 0.958\\
    Verma et al. \cite{verma2018diversity} & $\times$ & $\times$ & 0.784\\
    BIER \cite{opitz2017bier} & $\times$ & $\times$ & 0.952\\
    % UTH \cite{huang2017unsupervised} & $\times$ & $\times$ & 0.233\\
    HDC \cite{yuan2017hard} & $\times$ & $\times$ & 0.890\\
    A-BIER \cite{opitz2018deep} & $\times$ & $\times$ & 0.969\\
    ABE-8 \cite{Kim_2018_ECCV} & $\times$ & $\times$ & 0.979\\
    FastAP \cite{Cakir_2019_CVPR} & $\times$ & $\times$ & $\mathbf{0.985}$\\
    FashionNet \cite{liu2016deepfashion} & $\checkmark$ & $\checkmark$ & 0.764\\
    \hline
    %Our Baseline & $\times$ & $\times$ & 0.975\\
    %Our Bilinear Model & $\checkmark$ & $\checkmark$ & 0.981\\
    Our AHBN Model & $\checkmark$ & $\checkmark$ & 0.980\\
  \bottomrule
\end{tabular}}
\end{table}

\begin{figure}
\centering
\resizebox{0.48\textwidth}{!}{
\includegraphics[width=\linewidth]{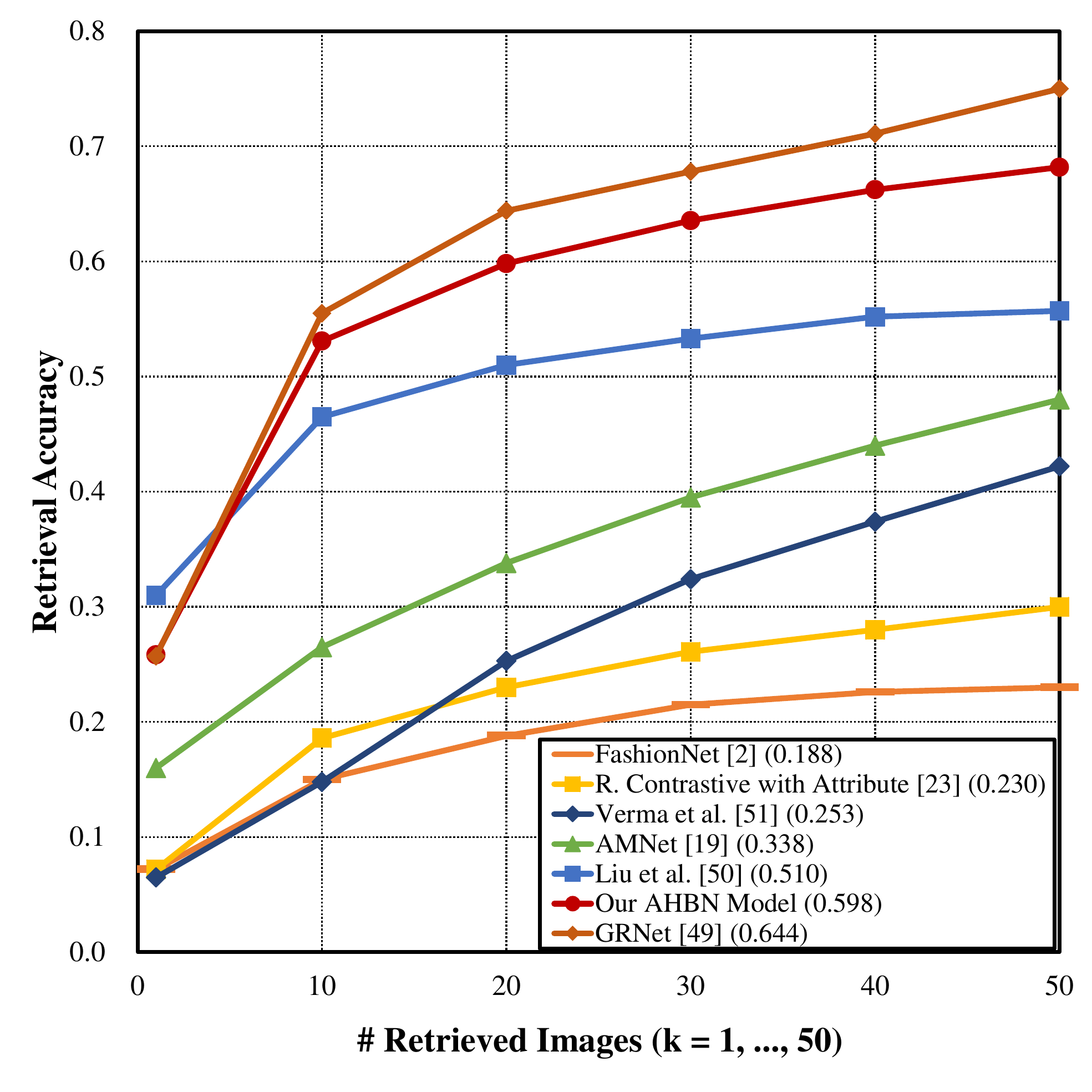}
}
%\vspace{-1em}
% \setlength{\abovecaptionskip}{-3pt}
% \setlength{\belowcaptionskip}{-5pt}
\caption{Top-$k$ matching results of compared models on DeepFashion Consumer-to-Shop benchmark \cite{liu2016deepfashion}}
\label{fig:result}
\end{figure}

\begin{figure*}
\centering
\resizebox{0.95\textwidth}{!}{
\includegraphics[width=\linewidth]{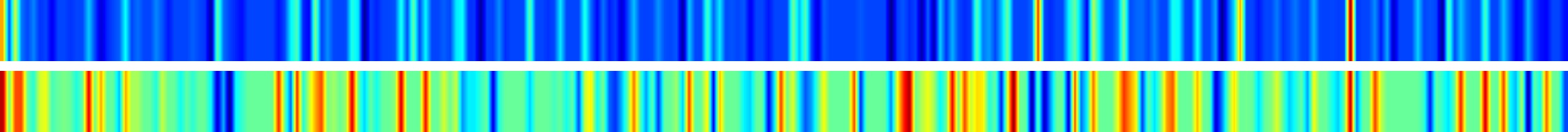}
}
%\vspace{-1em}
% \setlength{\abovecaptionskip}{-3pt}
% \setlength{\belowcaptionskip}{-5pt}
\caption{The attention weights obtained by softmax and sigmoid. The first row shows the weights obtained by softmax and the second row shows the weights obtained by sigmoid}
\label{fig:heatmap}
\end{figure*}

From the experimental results shown in Table~\ref{tab:ablation}, we obtain several observations as follows.
1) The {\bf Two-Branch w. $256$LM} model significantly outperforms the {\bf Single-Branch} model, showing the effectiveness of our heterogeneous two-branch design. 
2) The {\bf Single-Branch + ATTR} has better performance compared to the {\bf Single-Branch + LM}, which demonstates that the attribute branch contributes more in our AHBN model.
3) The {\bf Two-Branch w. $256$LM + Cat}, {\bf Two-Branch w. $256$LM + Mul} and {\bf Two-Branch w. $256$LM + Sum} indicate that the CBP~\cite{gao2016compact} can extract better feature representations.
4) The {\bf Two-Branch w. $256$LM} model also performs better than the {\bf Two-Branch w. $8$LM} model. We conjecture that the $256$-channel feature maps provide more useful information than the final $8$-channel heatmaps, as the former may contain localization cues for some latent object parts.   
5) Our {\bf AHBN} model achieves better results than the model without any attention module ({\bf Two-Branch w. $256$LM}) 
or the model with two separated attention modules ({\bf Two-Branch w. $256$LM + Sepa. Atten.}), 
which indicates that modeling the mutual interaction of the two branches is beneficial for estimating the importance of feature channels of both branches.
6) {\bf Two-Branch w. $256$LM} employs compact bilinear pooling after our two-branch network and {\bf Two-Branch w. $256$LM + BP} replaces the compact bilinear pooling by the standard bilinear pooling network. 
It is shown that the compact bilinear pooling has better performance than the traditional bilinear pooling.
7) We study the impact of the standard spatial attention mechanism by comparing {\bf Single-Branch} with {\bf Single-Branch + Spatial Atten.}, and find that adding spatial attention incur slightly worse performance. It shows that channel-wise attention has larger impact than spatial attention.
8) The performance of {\bf Our AHBN Model + $4\times4$} whose H and W setting are both $4$ is worse than {\bf Our AHBN Model}. The reason for worse performance may be the information loss. The performance of {\bf Our AHBN Model + $16\times16$} whose H and W setting are both $16$ is also inferior to {\bf Our AHBN Model} and the misalignment without pooling layers may be the main cause.
9) Our {\bf AHBN} model achieves better results than the model {\bf Our AHBN Model + Softmax}. We also visualize the attention weights obtained by softmax and sigmoid respectively in Figure~\ref{fig:heatmap}.
Due to the mutual exclusive nature of the softmax function, the softmax function generates a much more sparser attention weights than that with sigmoid. We suspect this over-sparsity may lead to information loss and consequently end up with worse performance.

\begin{figure}[t]
\centering
\resizebox{0.48\textwidth}{!}{
\includegraphics[width=\linewidth]{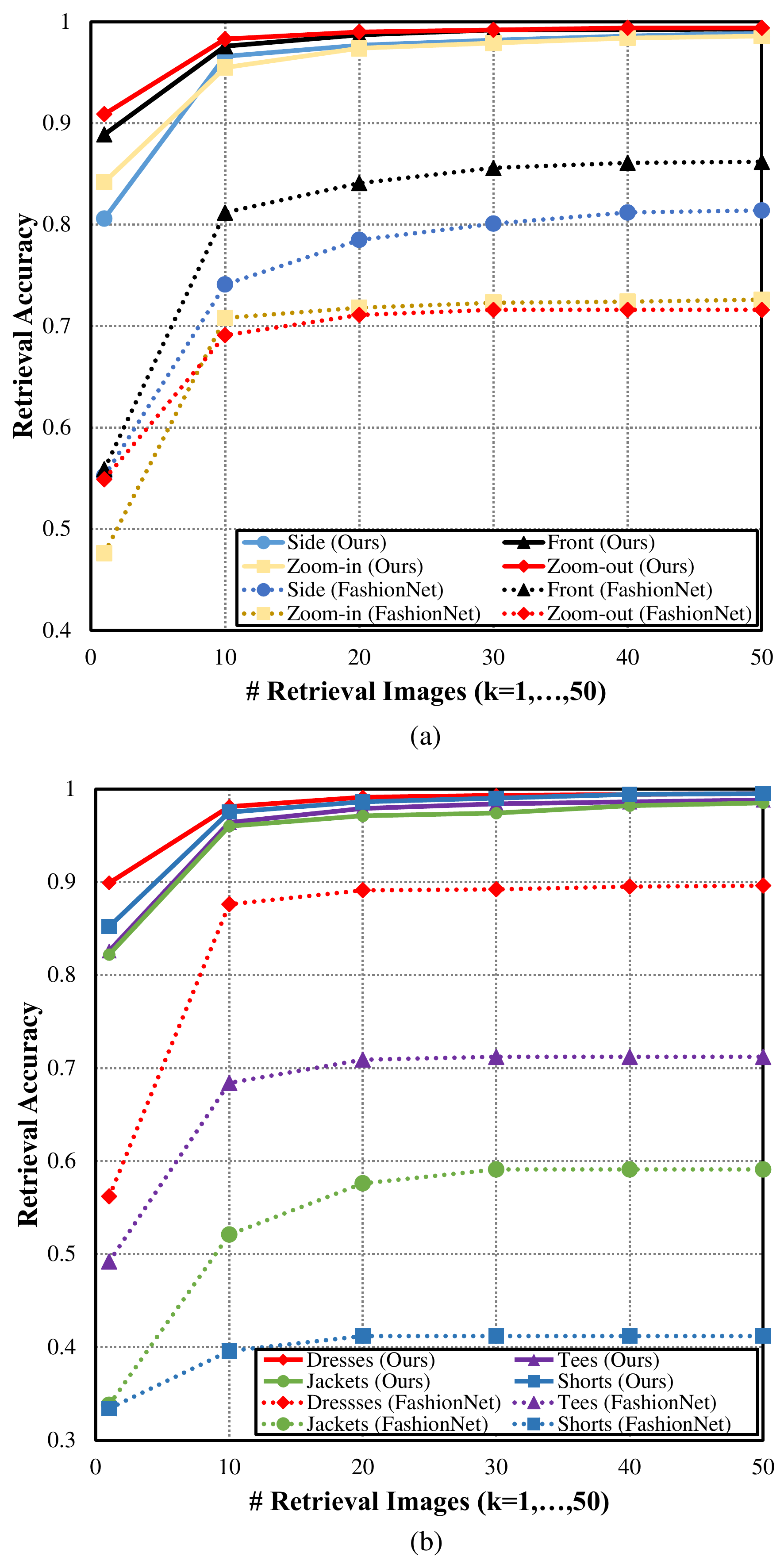}
}
%\vspace{-1em}
% \setlength{\abovecaptionskip}{0pt}
% \setlength{\belowcaptionskip}{-10pt}
\caption{Top-$k$ retrieval accuracies on DeepFashion Inshop benchmark~\cite{liu2016deepfashion} for different (a) poses and (b) categories.}
\label{fig:pose_category}
\end{figure}

\subsection{Comparison with State-of-the-arts}

In this section, we compare our proposed model with state-of-the-art approaches on three public benchmarks for fashion product retrieval. 
%Detailed performance of different models on different datasets is shown in Table~\ref{model:accuracy} and Table~\ref{modelonWTBI:accuracy}.

\noindent{\bf Exact Street2Shop.} 
Table~\ref{modelonWTBI:accuracy} lists top-$20$ retrieval accuracies on the six evaluated categories in the Exact Street2Shop dataset, including dresses, leggings, outerwear, pants, skirts and tops.
Our method performs better than others on all the six categories by a large margin.
Most evaluated algorithms perform better on ``Dresses'' and ``Skirts'' and worse on ``Leggings'' and ``Pants''. The reason may be that: there are a large variety of designs for Dresses and Skirts and they usually have more significant fashion symbols that can be used to distinguish one specific type from others; while the designs for Leggings and Pants are relatively not that diverse, which leads to a smaller inter-class difference. Because of the above reason, the fashion retrieval tasks for ``Dresses'' and ``Skirts'' are relatively easier than those for ``Leggings'' and ``Pants''.

\noindent{\bf DeepFashion Consumer-to-Shop Benchmark.} 
As shown in Figure~\ref{fig:result} and Table~\ref{model:accuracy}, our model performs better than all the compared methods except GRNet \cite{Kuang_2019_ICCV}. Note that the contributions of GRNet and ours are orthogonal. We can employ GRNet to improve our model furthermore.
% Note that FashionNet~\cite{liu2016deepfashion} also employs attribute and landmark annotations, but its performance is significantly worse than ours.
Compared to FashionNet, we use a more systematic way to model the interactions between the attribute and landmark branches.

\noindent{\bf DeepFashion Inshop Benchmark.} 
Different from Consumer-to-Shop, all images in this dataset are from the same domain. 
% So the evaluated retrieval performance is also better than that for Consumer-to-Shop.
As shown in Table~\ref{inshopmodel:accuracy}, our approach achieves the nearly best top-$20$ accuracy of $0.980$, slightly below the performance of FastAP \cite{Cakir_2019_CVPR}.
We also evaluate retrieval accuracies for different poses and clothes categories (see Figure~\ref{fig:pose_category}).
Our approach surpasses FashionNet by a large margin.

As shown in Table III and II, GRNet~\cite{Kuang_2019_ICCV} has a better performance on DeepFashion Consumer-to-Shop Benchmark. However, note that we surpass it on Exact Street2Shop. In Table IV, our performance is close to FastAP~\cite{Cakir_2019_CVPR}. 
GRNet proposed a Similarity Pyramid network which learns similarities between a query and a gallery cloth by using both global and local representations at different local clothing regions and scales based on a graph convolutional neural network.
FastAP employed a novel solution, i.e., an efficient quantization-based approximation and a design for stochastic gradient descent, to optimize average precision.
We believe that the contributions of GRNet, FastAP and ours are orthogonal. We will learn from their strengths to  improve our model furthermore.

Note that, for DeepFashion Consumer-to-Shop and Exact Street2Shop, the image in the query and gallery sets are from two different domains. In contrast, the query and gallery images in DeepFashion In-Shop are from the same domain. The cross-domain task is more difficult than the in-domain task, so the performance on DeepFashion Consumer-to-Shop and Exact Street2Shop datasets is significantly worse than that on DeepFashion In-Shop.
% \noindent{\bf Exact Street2Shop.} 
% Table~\ref{modelonWTBI:accuracy} lists top-$20$ retrieval accuracies on the six evaluated categories in the Exact Street2Shop dataset, including dresses, leggings, outerwear, pants, skirts and tops.
% Our method performs better than others on all the six categories by a large margin.

In summary, our proposed AHBN model achieves satisfactory retrieval performance on all the three benchmarks.

\section{Conclusion}
%In this work, we propose an attentional heterogeneous bilinear network for fashion image retrieval. The adopted two heterogeneous branches, \ie, an attribute classification network and a landmark prediction network, extract visual attribute information and object-part localization information. In certain sense, our model mimics the hypothesized two-stream visual processing system of human brain~\cite{goodale1992separate}. These two mutually complementary and interactive features are further integrated by an attentional bilinear pooling module. Note that each dimension of the final image representation encodes both attribute and localization information. The superior performance of our model is validated by our thorough experiments.
In this work, we propose an attentional heterogeneous bilinear network for fashion image retrieval. Compared to previous works, we introduce the localization information, which is extracted by a landmark network, to get a semantically rich second order feature by a bilinear pooling for each image. The localization information strengthens feature learning of key parts and minimizes distractions effectively. We also propose a mutually guided channel-wise attention to suppress the unimportant layers in consideration of localization and attribute. The superior performance of our model is validated by our thorough experiments.

However, there leaves a lot to be improved in our algorithm. One of the limitation of our algorithm is that we rely on human annotations to pretrain the two branches. This limitation prevents us from using massive unlabelled data. Recently, contrastive unsupervised representation learning~\cite{chen2020simple} has achieved significantly improved performance. For future work, we can incorporate unsupervised learning algorithms to pretrain the two branches in our framework and thus reduce the requirement on the labelled data.

\bibliography{aaai20}
\bibliographystyle{IEEEtran}
\end{document}